# Understanding Emails and Drafting Responses

# An Approach Using GPT-3

Jonas Thiergart, Stefan Huber, Thomas Übellacker

University College Maastricht

2021-02-04

# Introduction

Email is an essential medium for digital communication. By today, more than 300 billion emails are being sent and received daily. Although new communication means are continuously developed, the email-market is still predicted to grow to more than 360 billion emails and 4.48 billion users by 2024 (Clement, 2020).

Nevertheless, day-to-day email management is a flawed process. First of all, it takes time: professionals spend some 2.6 hours a day on checking, reading, responding to and writing emails (Plummer, 2019), which implies that the average knowledge worker "spends an estimated 28 per cent of the workweek managing email" (McKinsey & Company, 2012, para. 4). Secondly, it is inefficient: one study found that among professionals, "over 50% of messages received . . . on mobile email devices were not directly addressed to the recipient" (Mazmanian et al., 2013, p.7). Moreover, in this study, "for every eight work-related email messages participants received on mobile email devices . . . they replied to, or initiated, only one email message" (p.7). Other arguments for the inefficiency of email clients exist, but frankly, no one seems to dispute this. A survey on email customer service interactions found that "66% of U.S. consumers say the most important thing a brand can do is value their time" (Sacks, 2019, para. 1). Email communication in its current form squanders vast quantities of what is arguably the most precious resource. Moreover, aside from being a waste of human potential, labour time spent on routine email management, especially that of highly trained professionals, engenders high opportunity costs from a business perspective.

Innovation in Natural Language Processing (NLP) technology could be the tool to fix this problem. The recently released NLP technology known as GPT-3, in particular, might expand the possibilities for automating email management tasks. In this research paper, we draw on both technological and economic knowledge to explore these.

We ultimately seek to answer the question: Can one economically viably apply GPT-3 to understanding incoming emails and rationalising human effort in responding to them? To that end, we first set the stage by providing a general overview of the key terms and principles behind NLP and specifically, GPT-3, as well as their properties. Next, we argue for the thesis that it is possible and economically viable to use GPT-3 to rationalise specific email communication processes. Our argumentation works by substantiating three premises that, together, prove the thesis. First, we determine whether GPT-3 can answer emails with sufficient quality to augment human effort. To that end, we discuss GPT-3's capabilities and propose a technical way to do so. Second, we consider challenges relating to this application, like GPT-3's lack of access to business-internal knowledge, bias and its inaccuracy; we aim to prove that one can sufficiently address them. Third, we consider such a product's scalability and potential users to determine commercial viability. Finally, we synthesise our findings to prove the thesis while touching on our work's limitations.

# Background Information

## Natural Language Processing as an Answer to Complex Linguistics

Human language, also known as natural language, works differently from computer language. Natural language consists of many rules, exceptions, and particularities. Moreover, it is ambiguous and can vary in meaning depending on the context. Therefore, computer systems struggle to deal with it (Knight, 2016).

The discipline of Natural Language Processing (NLP) seeks to empower them to do so. NLP "is a collective term referring to automatic computational processing of human languages. It includes both algorithms that take human-produced text as input and algorithms that produce natural looking text as outputs" (Bahja, 2020, p.2). In other words, by combining disciplines like linguistics, statistics, data science and artificial intelligence, NLP engineers aim to model natural language in a way that allows computers to understand the content (natural language understanding) and express its own concepts (natural language generation) in natural language (Liddy, 2001).

Over the years, NLP technology has undergone several evolutionary stages. The earliest NLP models were "simple" mathematical representations of words and their meaning that computers could deal with (Pennington et al., 2014). The technology advanced with the rise of neural networks, allowing the creation of far more sophisticated models and the implementation of context-information (Dai & Le, 2015). Except for a few linguistically peculiar fields like biomedicine, where domain-specific language models seem to keep outperforming their non-specific counterparts (Poon & Gao, 2020), the current, state-of-the-art NLP models in most fields are "task agnostic" models that are fine-tuned on a specific task. This means the models are first trained on large, general datasets and then customised for particular tasks with another, smaller dataset (Colin et al., 2019). While this makes them high-performing, Brown et al. (2020), the creators of GPT-3, point out that in many cases, fine-tuning datasets either lack quality or are not available at all. Thus, the latest machine learning models were trained by increasingly large natural language datasets, resulting in cross-domain language models.

NLP already creates enormous value for the economy in various areas where large amounts of text data need to be understood, inferences drawn, and responses drafted. For example, NLP augments financial analysts by making predictions about the stock market based on analysing written "data from internet, news, blogs and social networking sites" (Bahja, 2020, p.5), or allows doctors to consider text data about past cases of illnesses, with one test improving cancer diagnosis accuracy by 22.6% (p.6). Moreover, NLP can summarise hard-to-read texts that are "full of abbreviations" or "spelling mistakes" and makes customer service chatbots possible (p.6). This has made NLP's presence felt, if not always consciously, in most peoples' daily lives.

## Generative Pre-Trained Transformer 3 Does Not Require Training Data

Generative Pre-trained Transformer 3 (GPT-3), as a "task-agnostic" language model, could make possible even more applications of NLP. The system, which the research lab OpenAI released for testing in August 2020, can compete with previous, fine-tuned models for many use cases and on many benchmarks without any fine-tuning. According to its creators, it is pre-trained on a dataset over ten times larger than that of any comparable models that includes i.a. Wikipedia and a filtered version of the Common Crawl dataset (Brown et al., 2020). OpenAI has set up an application programming interface (API) where users access GPT-3 and "program" it by entering words in human language, giving instructions on what to do, and, optionally, a few examples of the desired output. That way, GPT-3 gets a similar number of instructions to what humans would get when asked to perform a task (unlike older NLP models, which require large sets of training data), providing the flexibility needed for a wide range of applications. As such a general-purpose language model, GPT-3 can be applied to a wide variety of tasks. GPT-3 has proven capable of generating poetry or articles indistinguishable from human authors, computer code for web interfaces, and product or job descriptions (Dale, 2020).

# Technical Viability

Having explained relevant concepts, in the following, we evaluate the technical viability of using GPT-3 to understand emails and generate responses. To do so, we identify the individual steps required and determine that they can be performed.

## Understanding Incoming Emails

### Context Understanding Using Text Classification

The very first step in understanding an email might be to classify it. Classification is used to distinguish spam from relevant messages and help to understand the context of an email (Wasi et al., 2015, p. 130). In many businesses, it makes sense to classify emails to be more efficient in responding to them. Consider, for example, a company's support email system. Incoming emails could be classified into predefined categories to prioritise and distribute them among different persons or departments. Moreover, email providers like Gmail, for instance, classify consumer emails into categories like "social", "promotions", "updates" and "forums" (Izatt, 2020) by default. A more sophisticated system may allow for using template responses or automatically triggering other business processes based on an incoming email classification.

Common techniques for text classification range from low-level semantic analysis to more sophisticated methods like deep learning. A simple approach would be using a Naive Bayes Classifier considering words in the email's body and title, assuming their order is irrelevant (Al-Alwani, 2014, p. 691). Still, a comparison of different supervised learning algorithms shows that other algorithms easily outperform Naive Bayes Classification (Caruana & Niculescu-Mizil, 2006, p. 165). Those may perform well at rationalising email classification processes, but only if ample training data is available.

The currency of GPT-3 implies academic literature on its uses for text classification is limited, so Übellacker (2021) conducted empirical testing, concluding that GPT-3's capabilities are sufficient in this regard. As previously mentioned, prompts in the GPT-3 API are formulated similarly to commands that one would make to a person. As a result, providing GPT-3 with a list of possible categories, along with a brief description of the email itself and what it should do (without giving specific examples) already works well. GPT-3 understands categories just by their titles, adding to the algorithm's flexibility and enabling applications where no or insufficient training data is available. Depending on the email body's length, the GPT-3 prompt is also relatively short (usually 110-300 tokens, a unit for text length), which means the classification is relatively fast (Übellacker, 2021). Hence, classifying texts with GPT-3 works (especially when there is no training data available). Nonetheless, it is more expensive and inaccurate than text classification using fine-tuned models. Hence, if there is training data available, we recommend using those. However, due to flexibility reasons, in the course of the paper, we follow the classification approach with GPT-3. The classification allows for assessing an email's context and further processing it.

**Extracting Relevant Information from Email Body**

Another vital step in understanding emails is extracting information for further processing. There are several methods for extracting information from unstructured data (in this case, email bodies in plain text), including Named-Entity Recognition (NER). NER systems are used to extract named entities like organisations, geographical locations, person names or dates from natural texts (Singh, 2018). If we examined the context-category of an email, in the first step of using text classification, we can now use information extraction to find category-specifically relevant information. In case of an email invite to an event, for example, NER could be used to gather event details that can then automatically be added to a calendar. Another application would be an incoming order email, where NER then is used to extract information about the articles that are being ordered. NER technologies like Duckling (Wit.ai, 2020) and spaCy (Schmitt et al., 2019) are pre-trained to be capable of recognising certain entities like dates, numbers, distances, emails, persons, organisations, etc. They can also be trained to detect custom entities when enough training data is available (Nazakat, 2020).

In contrast, GPT-3, being pre-trained, can extract named-entities with little or no training data. Test prompts show that GPT-3 can extract event information without any specific customization (Übellacker, 2021). This information could easily be forwarded to a user's calendar system, further processed or used later for email response generation. Another example shows that it is possible to extract detailed order information from an incoming email that could potentially be automatically forwarded to a customer relations system (Übellacker, 2021).

## Generating Responses With GPT-3

After email classification and information extraction, the last step toward rationalizing email communication is generating a response text, which is something GPT-3 can do. Historically, natural language generation technologies suffered from the chicken-and-egg problem of needing a large, annotated data set to train models, but there being a lack of people to write and annotate text. Fortunately, GPT-3 offers new possibilities and is innately capable of composing text in cross-domain applications (Brown et al. 2020). In the context of email automation, GPT-3 can be used to answer emails, feeding the prompt with background information gathered during the email understanding process and the corresponding email thread.

By all indications, the quality of GPT-3 generated text is many ways similar to that of text written by humans. GPT-3 can generate grammatically correct, coherent responses. With its vast, previous knowledge, it can even answer emails where the message is not completely specified, like open-ended questions (Dale, 2020). This means that, for example, if a user asks

GPT-3 for a nice restaurant in a given city, it will be able to recommend a place to eat. Even more remarkably, it can do so in a conversational manner and mimic empathy (Aronsson et al., 2020), allowing the email recipient to feel understood.

Nevertheless, GPT-3 sometimes interprets messages awkwardly. It is capable of answering general emails but has trouble in specific cases. For example, if it is asked something that requires information that is not in the training dataset, such as internal business data, it cannot respond appropriately. However, the GitHub repository by Übellacker (2021) shows that GPT-3 can achieve impressive results when given enough context. This repository does not show exhaustive testing with different inputs (which calls for further research), but it nevertheless seems likely that GPT-3 can be used in a way that allows it to understand text with great accuracy.

# Solving Main Limitations

In the previous chapter, we showed that using GPT-3 to understand emails and respond to them works in theory. However, this approach has inherent, technical limitations that could make it unworkable in practice. In this section, we argue that these can be addressed.

## GPT-3 Makes Mistakes

One of the main limitations of GPT-3 is its lack of reliability. In some cases, especially as a result of weak instructions, the produced output is wrong content-wise, reflects undesirable biases from the training data such as racism (Floridi & Chiriatti, 2020) or is simply nonsense (Dale, 2020). There are specific, fine-tuned models capable of detecting such issues (Al-Hassan & Al-Dossari, 2019) in GPT-3's output. To that end, it is advisable to implement additional software in the architecture. Nevertheless, these measures leave room for mistakes.

Thus, it becomes clear that GPT-3 is currently not made for automation. Instead, we propose retaining a "human-in-the-loop": The solution should be understood as a tool to augment people and support their email writing, rather than a replacement. This could be achieved by adding functionality to email clients that allows users to pick whether to fully accept, edit or discard automatically drafted emails.

## GPT-3 Is a Few-Shot-Learner

Another limitation relates to the fact that, as previously mentioned, GPT-3 works without large fine-tuning data sets. By its nature, the language model has knowledge about almost any domain. It learns based on only a few commands on what to do and ideally, some examples of the desired output (few-shot learning) (Wang et al., 2020). We describe above how GPT-3 therefore performs well in tasks that require only general information but struggles with tasks that require further knowledge. However, when writing emails in a work environment (which may include a large portion of emails), it is often necessary to use just such (business-internal) information that GPT-3 has never seen before. In the following, we propose a solution that allows us to augment GPT-3 with business-internal information.

To determine how to include business-internal information in emails, one must first understand that it comes in a variety of forms. In many cases, information is stored in enterprise resource planning (ERP) or customer relationship management (CRM) systems, wikis, email archives or product documentations. Moreover, recent years show a steady increase in unstructured data, making them less accessible for software systems. While structured data like tables and databases are often easier to implement, GPT-3 might also want to consider information from PDFs, images and video files (Harvard Business Review, 2020).

The main obstacle in the way of augmenting GPT-3 with business-internal information is that since we cannot feed the GPT-3 prompt with a lot of data (OpenAI, 2020), it needs to be provided with precisely the right data. To address this, we propose a software architecture that allows GPT-3 to analyse an email's content and evaluate which information topics are most relevant for responding to it. Thus, we need a solution to search through large amounts of unstructured resources and return text information to questions.

Cloud service providers offer different AI services that could perform this task. Current cloud search services that search over large amounts of unstructured business data with keyword searches are not based on keyword-matching, but they understand the content and the context to deliver precise search results (Harvard Business Review, 2020; Talia, 2013). Most services like Kendra by Amazon Web Services (AWS) or IBM's Watson Discovery return paragraphs that are likely to contain the answer to a given question or given keywords. They are straightforward when adapting the product for new customers and have many connectors to different data sources such as SharePoint or Google Drive (which becomes relevant when companies use cloud services to store their data). From the perspective of economic viability, Watson Discovery is the most affordable option for this type of search (Robinson 2020; Amazon 2020; IBM 2020).

Azure Cognitive Search, the search service of Microsoft, is in a similar price range as Watson Discovery, but it works differently. Instead of a single service that returns passages to searches, Azure's approach to searching through unstructured data envisions a stream of different services that are configured and connected in a row. In contrast to the previously analysed services, Azure's Cognitive Search itself only searches through structured data. Therefore, it is served as a final step of a series of other AI models used to prepare information, rather than as an independent service.

Engineers in the discipline of knowledge mining develop approaches that aim to get insights out of this great variety of unstructured information. To that end, knowledge mining applies pre-built AI-services such as computer vision, sentiment analysis or language recognition to extracting key information out of these different types of data and makes it accessible as structured data (Harvard Business Review, 2020). Besides the option for custom models, Azure offers predefined knowledge mining models such as computer vision and layout understanding that enrich the data, i.e. gain structured information from unstructured content. Thereby Cognitive Search allows searching through a range of unstructured input such as images, audio files and even videos, which goes far beyond the capabilities of search services of other providers (Microsoft, 2020).

Consequently, while Watson Discovery might be a good fit for companies that only want GPT-3 to consider information from text documents, Azure search with its affiliated functions (Robinson, 2020) currently appears to be a workable as well as the most suitable tool for providing GPT-3 with the business-internal information necessary for drafting email responses.

# Economic Viability

Having demonstrated that applying a GPT-3-based software to rationalising email understanding as well as drafting is feasible and that technical solutions exist for the main challenges, the next step is to determine whether deploying it at scale is economically viable. This depends on two types of factors: financial factors like the cost structure for the service provider and benefit to customers, as well as the demand side, the potential size of the market for this service. We devote a section to each of these concepts, beginning with the financial factors.

## Cost Analysis

Generally, software products are easily scalable because of their near-zero marginal cost (MC). The physical infrastructure like cloud services (the main cost factor for scaling) is becoming so cheap that one theorist coined the term "zero marginal cost society" (Rifkin, 2014) to describe the situation. This is because the effect of diseconomy of scale (which limits scalability) only takes place when the long-run average cost per unit (LRAC) increases with the quantity produced; a constant (in this case, near-zero) MC implies that LRAC cannot increase.

In the case of GPT-3-based email rationalisation, MC is not zero, but still constant, meaning such a service would be easy to scale (provided customers were prepared to pay a price above MC for it). The reason for the non-zero MC is that GPT-3's creators' business model was to receive payment proportionally to the amount of text processed by the GPT-3 software. This payment will be on the order of magnitude of 6 dollar cents for 1000 "tokens" of text (MLK, 2020). If this were the only MC, the average cost per unit would merely approach MC in the long run, leading to no increase in the LRAC and ergo, no diseconomy of scale. However, other costs of providing the service arise: the "customisation costs" of adapting the software to a new customer (integrating business internal data) - these may indeed increase LRAC if one starts with customers with low customisation cost. Overall, the product may, therefore, not necessarily be scalable to fill the whole market demand. Still, it appears the MC of the product would initially be constant (there are many similar customers, as we will show below) - possibly making it scalable to a degree that would allow a provider to pay fixed costs (e.g., for labour and a flat rate for the search service), rendering the product profitable.

The earlier thoughts regarding scalability focused on proving there would be no (early) diseconomy of scale since there is no change in LRAC as quantity increases. It was still not clear whether adopting the technology is cost saving. To answer that question, we will make and compare rough estimates of the cost of generating a given length of text with GPT-3 or writing it manually. For the sake of argument, we will consider an email with a length of 500 words. Considering human adults can read 300-575 words per minute (Nelson, 2012), humans would need around 1 minute to read such an input text as well as a likely longer, but hard-to-estimate time to write a response. Assuming workers reading this text were paid the California minimum wage of 15 dollars/hour (US Department of Labor, 2021), the wage cost for merely reading it would be around 25 cents. The total cost of responding would be significantly higher. To estimate the cost of responding with GPT-3, one needs to estimate the total number (and therefore price) of tokens required to generate a response in addition to the wage cost of manually editing out any inaccuracies before sending. We know "2048 tokens translated to words will be ~ 1500 words" (He, 2020, para. 3), implying that a 500-word email will be around 700 tokens long. The total number of tokens necessary for generating an email response is greater than the length of the email itself because of how GPT-3 functions. To create a response email using the business information connector, one needs to feed text into GPT-3 several times, which, based on our proposed setup, totals a fixed 63 tokens plus two times the length of the preceding email and one time the length of the generated response R (Übellacker, 2021). This works out to tokens. At 6 cents/1000 tokens, the GPT-3 cost is at least about 8.8 cents. If one assumes the response is as long (700 tokens) as the received email, the total GPT-3 cost increases to about 13 cents. Therefore, while the cost of manually responding is 25 cents plus the wage cost of writing a response, the cost of using GPT-3 is 13 cents plus the wage cost of editing a response. Assuming that editing a reply takes less time than writing one from scratch (or that wage costs are usually above minimum wage), the difference becomes even greater. As a result, the effective cost of manually responding to emails is higher than the constant, marginal cost of doing so with GPT-3. Therefore, applying a GPT-3-based software at scale would save users time and money even at a price well above MC and therefore be economically viable for providers.

## Market Demand

Aside from thinking about the costs associated with using GPT-3 to rationalise email communication, one must identify the potential demand to accurately judge whether it is an economically viable solution. To do so, one must define some criteria users need to fulfil; in this case, what type of entity they are, as well as how many and what type of emails they send in their value creation chain.

We know that around 28% of knowledge workers' work hours are devoted to email management (McKinsey & Company, 2012). It seems reasonable to assume that most email management happens at work and that employers generally provide employees with work equipment. Hence, the natural customers for the described product appear to be employers, i.e., private firms or the government. Therefore, the answer to the question of the type of entity of users seems to be clear.

The more interesting question is that of the nature and role of emails in their value creation chain. From the background information section, we know that some fields involve a linguistic complexity that general language models like GPT-3 "understand" poorly. On the other hand, certain types of messages, like advertisements, do not need to be generated; they are obviously reused verbatim. Therefore, the subject matter to be understood and responded to must neither be overly complex nor too simple. Another issue is that of importance - at least until the technology becomes more reliable, firms might prefer to draft certain highly consequential documents like deal proposals manually. Finally, because the marginal costs of the software include the per-volume cost demanded by OpenAI for the use of the GPT-3 API, profits and pricing depend on the quantity of text understood or generated, signifying that firms with a high volume of email traffic are preferable customers. Therefore, the ideal users would be corporate or public employers whose work processes involve understanding and responding to large volumes of "semi-complicated" text where the possible damage of a misunderstanding is not excessively high.

What types of firms, then, fulfil these criteria? Our research indicates that a market for GPT-3-based email rationalisation exists in several different sectors of the economy, of which we shall explore just a few. In all sectors, the damage of a small mistake in wording seems minor as content generally involves neither vast amounts of money nor human safety.

The insurance industry, where correspondence consists primarily in handling "insurance claims and policy adjustments" (Stoeckli et al. 2018, p.299), represents one example. Studies of this sector have found that even in the digital world, direct communication between two people results in a significant increase in trust and convincement (Maas & Bühler, 2015, p.22), implying that completely automating insurance customer service is not desirable. Simultaneously, response speed and accessibility for requests are considered two critical wishes of customers (p.22), which has led to a shift from offline media like call centres to online media like email or apps at all stages of the customer journey (p.22). Moreover, surveyed experts estimate the automation potential at 28% of the industry's value creation (p.38). These combined trends signify that GPT-3-based email customer service with a "human-in-the-loop" could meet many of the insurance sector's future demands. Although insurance markets differ in the frequency of customer interaction, with e.g., healthcare usually requiring more customer service than life insurance (Stoeckli et al. 2018, p.299), the volume of communication is likely enormous, as health insurance coverage alone is near-universal in many countries (New York State, 2011).

The market for utilities such as energy faces similar trends. Analyses show that "administrative processes in customer management and billing (including changes in provider, address, or product) are proliferating. Distributed generation and multiple channels are resulting in more convoluted and error-prone processes" (Peters et al., 2016). At the same time, it is becoming clear that "traditional energy retailers need to adapt to the digital age—fast. Unencumbered by large sales forces and expensive call centres, challengers keep costs low by communicating with customers primarily online" (Lehrke et al., 2018). Again, the combined developments of increasing administrative processes and a shift toward online communication imply that the utilities sector's customer service might present a market for rationalised email communication. The volume of traffic is considerable, seeing as utilities, perhaps more so than health insurance, is a near-universal market in terms of number of customers.

A third market for GPT-3-based email rationalisation may be that of email communication in public administration. For example, the German government recently introduced a secure email client known as De-mail, designed to bundle all communication between citizens or private firms and the federal government's institutions into a single, digital channel. This includes applications for transfer payments like maternity support, customs clearances, the information provided for court proceedings, answers to general questions, etc. (Bundesverwaltungsamt, 2018). As with the commercial firms, the government responds to the demand of citizens for practical communication and cost incentives by shifting the bulk of communication to online media like email. The conditions for using GPT-3 with a human-in-the-loop are met; there exists a large volume of not completely standardisable email communication. The general point applies more broadly, as, according to the latest UN E-Government Survey, the number of countries that provide government services via email or apps has significantly risen in all sectors (UN Department of Economic and Social Affairs, 2020).

These three examples indicate that a considerable market for the type of rationalised email communication delivered by GPT-3-based software exists. This, combined with the cost-effectiveness of the technology shown earlier, suggests that the already low fixed costs of developing the software would become negligible if the product were scaled to fill even a fraction of the potential market size. Ergo, this technology would probably be economically viable.

# Conclusion

In this paper, we explored the possibility of rationalising email communication using GPT-3. Primo, we demonstrated the technical feasibility of understanding incoming emails and generating responses, drawing on literature from the disciplines of software engineering as well as data science. We found that it is indeed possible to use GPT-3 to classify emails, apply named-entity recognition and ultimately to generate emails. Secondo, we applied knowledge from both business studies and, again, software engineering to identify ways to tackle challenges we encountered. We proved that company internal information can be implemented without conventional fine-tuning. Moreover, weaknesses inherent in GPT-3 such as bias and the likeliness of mistakes can be tackled by means of an appropriate implementation of a human-in-the-loop. Terzo, we argued for the economic viability of such a solution by analysing costs and market demand, utilizing common economic principles as well as market research literature. To that end, we showed that the marginal cost is relatively constant (leading to an absence of the effect of diseconomy of scale, indicating scalability), estimated that the technology would be cost saving compared to manual email drafting and finally showed that even given its limitations, a large demand for such a solution exists in fields from customer service in insurance or utilities to public administration. We conclude that applying GPT-3 to rationalising email communication is feasible both technically and economically, assuming the particular architecture we propose. Consequently, this idea might be promising for entrepreneurs.

However, we realise our findings are limited. For instance, we cannot know whether workers would accept this solution or that the time spent editing responses is far below that spent writing them. Furthermore, a major limitation results from the customization costs: In our proposed solution, GPT-3 can access company internal knowledge by making use of cognitive search services. Depending on the information structure of a company, the costs of making this knowledge available to search services might surpass any possible savings. To answer such questions, one would need to conduct further, experimental testing. Another caveat is that the uses of GPT-3 prompts cited in this paper are not optimized and in need of further improvement. Moreover, in this study, we ignored that other, disruptive technologies might replace traditional email communication.

Despite its limitations, the study honed our understanding of the nature and scope of use cases of general language models in a business context. We proved the possibility of applying GPT-3, a pre-trained model, to dynamic real-life use-cases where previously, a lot of data was necessary to train a fine-tuned model. This realization could have significant implications on future business process design. We are optimistic that entrepreneurship and further research into this idea will reduce the inefficiency of email communication.

Business Review. https://hbr.org/2019/01/how-to-spend-way-less-time-on-email-every-day

Poon, H., & Gao, J. (2020, August 31). Domain-specific language model pretraining for biomedical natural language processing. *Microsoft Research*. https://www.microsoft.com/en-us/research/blog/domain-specific-language-model-pretraining-for-biomedical-natural-language-processing/

Rifkin, J. (2014). *The zero marginal cost society: The Internet of things, the collaborative Commons, and the eclipse of capitalism*. St. Martin's Press.

Robinson, S. (2020, April 9). Enterprise search software comparison. *SearchContentManagement*. https://searchcontentmanagement.techtarget.com/feature/Enterprise-search-software-comparison

Sacks, R. (2019, March 1). *Use natural language processing to automate customer support*. Medium. https://medium.com/ibm-watson/use-ibm-watson-natural-language-classifier-to-automate-customer-support-b35c2761211c

Sagar, R. (2020, June 3). *OpenAI Releases GPT-3, The Largest Model So Far*. https://analyticsindiamag.com/open-ai-gpt-3-language-model/

Sewain, A. (2020). *How NLP can increase efficiency by reducing time spent on emails*. 2021.AI. https://2021.ai/nlp-increase-efficiency/

Schmitt, X., Kubler, S., Robert, J., Papadakis, M., & LeTraon, Y. (2019). A Replicable comparison study of NER software: StanfordNLP, NLTK, OpenNLP, spacy, gate. *2019 Sixth International Conference on Social Networks Analysis, Management and Security (SNAMS)*. https://doi.org/10.1109/snams.2019.8931850

Singh, S. (2018, July 6). *Natural language processing for information extraction*. arXiv.org. arXiv:1807.02383
22